\pdfoutput=1
\documentclass{article}
\usepackage{ijcai22}

\usepackage{times}
\usepackage{soul}
\usepackage{url}
\usepackage[hidelinks]{hyperref}
\usepackage[utf8]{inputenc}
\usepackage[small]{caption}
\usepackage{graphicx}
\usepackage{amsmath}
\usepackage{amsthm}
\usepackage{booktabs}
\usepackage{algorithm}
\usepackage{algorithmic}
\usepackage{amsfonts,amssymb}
\usepackage{mathrsfs}
\usepackage{makecell}
\usepackage{color}
\usepackage{amsfonts,amssymb}
\usepackage{xspace}

\newcommand{\citet}[1]{\citeauthor{#1}~\shortcite{#1}}

\makeatletter
\def\@fnsymbol#1{\ensuremath{\ifcase#1\or \dagger\or *\or \ddagger\or
		\mathsection\or \mathparagraph\or \|\or **\or \dagger\dagger
		\or \ddagger\ddagger \else\@ctrerr\fi}}
\makeatother

\title{A Survey of Vision-Language Pre-Trained Models}

\author{
Yifan Du$^{1,3}$\footnote{Equal Contribution.}\and
Zikang Liu$^1$\footnotemark[1]\and
Junyi Li$^{1,2}$\And
Wayne Xin Zhao$^{1,3}$\footnote{Corresponding Author.}\\
\affiliations
$^1$Gaoling School of Artificial Intelligence, Renmin University of China\\
$^2$DIRO, Universit\'{e} de Montr\'{e}al\\
$^3$Beijing Key Laboratory of Big Data Management and Analysis Methods\\
\emails
\{yifandu1999, jasonlaw8121, batmanfly\}@gmail.com,
junyi.li@umontreal.ca
}

\begin{document}

\maketitle

\begin{abstract}
As transformer evolves, pre-trained models have advanced at a breakneck pace in recent years. They have dominated the mainstream techniques in natural language processing~(NLP) and computer vision~(CV). How to adapt pre-training to the field of Vision-and-Language~(V-L) learning and improve downstream task performance becomes a focus of multimodal learning. In this paper, we review the recent progress in Vision-Language Pre-Trained Models~(VL-PTMs). As the core content, we first briefly introduce several ways to encode raw images and texts to single-modal embeddings before pre-training. Then, we dive into the mainstream architectures of VL-PTMs in modeling the interaction between text and image representations. We further present widely-used pre-training tasks, and then we introduce some common downstream tasks. We finally conclude this paper and present some promising research directions. Our survey aims to provide researchers with synthesis and pointer to related research.
%Vision-language Pre-training~(VLP) aims to enhance model's ablity of understanding the relationshop between vision and language by pre-training on large-scale image-text dataset. Following the pre-training and fine-tuning paradigm, VLP models are able to overcome the lack of image-text data and achieve excellent performance on downstream vision-language task. In this paper, we present an overview of typical solutions and mainstream approachs in VLP area. First, we briefly review transformer and advances of pre-training method. We then introduce several ways to encode raw image-text data into single-modal embeddings. As the core content, we categorize several mainstream architectures of VLP models in detail and discuss the applicable scene of each category. We also present typical pre-training tasks and describe the detailed pre-training process. Then we introduce several commonly-used downstream task and how VLP models are adapted to them. Finally, we conclude the paper and present several future directions. 
\end{abstract}

\section{Introduction}\label{intro}
We now live in a world with various modalities~(voice, vision, odors, \emph{etc.}), among which vision and language are two critical ones. In academia, there exist large amounts of works focusing on V-L tasks. These tasks require the agent to jointly process information from these two modalities and utilize them to answer complex questions. For example, visual question answering~(VQA)~\cite{antol2015vqa} takes an image and the corresponding question as input and gives the correct answer; image captioning~\cite{lin2014microsoft} generates a description for a given image.

Deep learning has revolutionized the fields of artificial intelligence. Various deep models have been applied to solve V-L tasks, such as recurrent neural network (RNN)~\cite{arevalo2017gated}, convolutional neural network (CNN)~\cite{huang2020pixel} and transformer~\cite{vaswani2017attention}. Despite the success of deep learning, most of these models are designed for specific tasks, which leads to poor transferability. Pre-training a huge model on large-scale general datasets and then fine-tuning it on specific downstream tasks is one technique to increase transferability. Pre-training is first discovered to be effective in the field of CV~\cite{simonyan2014very}. After the proposal of transformer~\cite{vaswani2017attention} and BERT~\cite{devlin2018bert}, the paradigm of pre-training and fine-tuning becomes prevalent in the field of NLP. With its powerful capability to model long-range dependency, transformer has become the backbone of most Pre-trained Language Models~(PLMs). BERT and GPT-3~\cite{brown2020language} are typical PLMs that significantly outperform previous methods and achieve new state-of-the-art results on various downstream tasks. 

Due to the success of pre-trained models in the field of CV and NLP, many works have tried to pre-train large-scale models on both vision and language modalities, called Vision-Language Pre-Trained Models (VL-PTMs). By pre-training on large-scale image-text corpora, VL-PTMs can learn universal cross-modal representations, which are beneficial for achieving strong performance in downstream V-L tasks~\cite{zellers2019recognition,tan2019lxmert}. For example, LXMERT~\cite{tan2019lxmert} employs a dual-stream fusion encoder to learn V-L representations, significantly outperforming traditional models on VQA~\cite{antol2015vqa}, NLVR$^2$~\cite{suhr2018corpus} tasks via pre-training on 9.18M image-text pairs. Besides, VL-PTMs achieve strong results on many other V-L tasks like visual commonsense reasoning~\cite{zellers2019recognition} and image captioning~\cite{lin2014microsoft}. %It is first found to be useful in the field of CV. Various variants of CNN~\cite{he2016deep,simonyan2014very} are pre-trained on ImageNet~\cite{deng2009imagenet} and fine-tuned on downstream tasks like image classification~\cite{he2016deep}, object detection~\cite{sermanet2013overfeat}, etc. After the proposal of BERT~\cite{devlin2018bert}, the paradigm of pre-training and fine-tuning also prevails in the area of NLP. Considering the success of pre-training and fine-tuning on CV and NLP, it is natural to apply this paradigm on V-L tasks. There are some early works that first pre-train a V-L model on large amounts of (image, text) pairs and fine-tune it on downstream tasks like VQA, NLVR~\cite{suhr2018corpus}, IC, etc.~\cite{li2019visualbert,lu2019vilbert,zhou2020unified}
%With the advance of transformer architecture and the increasing of computation power, more and more V-L PTMs appears in the field of multimodal learning. 
These VL-PTMs utilize different single-modal encoders, elaborate V-L interaction schemes, and devise various pre-training tasks. 

However, there lacks a comprehensive survey to summarize the recent progress in this field. \citet{mogadala2021trends} mainly reviews existing V-L tasks, datasets, and traditional solutions but rarely introduces V-L pre-training methods. \citet{ruan2022survey} focus on Video-Language PTMs instead of Vision-Language PTMs. Different from them, our survey aims to present a thorough review of VL-PTMs, which summarizes recent research progress and provides pointers to related research. We present the recent mainstream VL-PTMs in Table~\ref{summarize}.

Basically, there are three steps to pre-train a VL-PTM: 1) encode images and texts into latent representations preserving their semantics~(Section~\ref{inputrep}); 2) design a performant architecture to model the interaction between two modalities~(Section~\ref{architecture}); and 3) devise effective pre-training tasks to train the VL-PTMs~(Section~\ref{pretraining}). After learning universal vision and language features, VL-PTMs can be fine-tuned on various downstream V-L tasks~(Section~\ref{downstream}). Finally, we conclude this survey and point out some promising research directions in Section~\ref{conclusion}.

\begin{table*}[!htb]
\small
\centering
\resizebox{1.0\textwidth}!{
\begin{tabular}{llllll}
\toprule
VL-PTM & Text encoder & Vision encoder  & Fusion scheme & Pre-training tasks & Multimodal datasets for pre-training\\
\midrule
\textbf{\small{Fusion Encoder}} \\
VisualBERT~\shortcite{li2019visualbert}  & BERT  & Faster R-CNN   & Single stream & MLM+ITM & COCO\\ 
%Unicoder-VL~\shortcite{li2020unicoder} & BERT  & Faster R-CNN   & Single stream & MLM+MRC+ITM\\
Uniter~\shortcite{chen2020uniter} & BERT  & Faster R-CNN   & Single stream & \makecell[l]{MLM+ITM+WRA+MRFR+MRC} & CC+COCO+VG+SBU\\ 
OSCAR~\shortcite{li2020oscar} & BERT  & Faster R-CNN   & Single stream & MLM+ITM & CC+COCO+SBU+Flickr30k+VQA\\
InterBert~\shortcite{lin2020interbert}  & BERT  & Faster R-CNN   & Single stream    & MLM+MRC+ITM & CC+COCO+SBU\\
ViLBERT~\shortcite{lu2019vilbert} & BERT  & Faster R-CNN   & Dual stream & MLM+MRC+ITM & CC\\ 
LXMERT~\shortcite{tan2019lxmert}     & BERT  & Faster R-CNN   & Dual stream & \makecell[l]{MLM+ITM+MRC+MRFR+VQA} & COCO+VG+VQA\\
% VL-BERT\shortcite{su2019vl}     & BERT  & Faster R-CNN+ ResNet   & Single stream & MLM+MRC & CC+BookCorpus+Wiki\\
VL-BERT~\shortcite{su2019vl}     & BERT  & Faster R-CNN+ ResNet   & Single stream & MLM+MRC & CC\\
Pixel-BERT~\shortcite{huang2020pixel}   & BERT  & ResNet   & Single stream & MLM+ITM & COCO+VG\\
Unified VLP~\shortcite{zhou2020unified} & UniLM  & Faster R-CNN   &  Single stream  & MLM+seq2seq LM & CC\\
% UNIMO~\shortcite{li2020unimo}     & BERT, RoBERTa  & Faster R-CNN   & Single stream & \makecell[l]{MLM+seq2seq LM+MRC+MRFR+CMCL} & BookCorpus+Wiki+OpenWeb~Text+OpenImages+COCO+CC+VG+SBU\\
UNIMO~\shortcite{li2020unimo}     & BERT, RoBERTa  & Faster R-CNN   & Single stream & \makecell[l]{MLM+seq2seq LM+MRC+MRFR+CMCL} & COCO+CC+VG+SBU\\
% E2E-VLP    & BERT  & ResNet   & Single stream \\
SOHO~\shortcite{huang2021seeing}     & BERT  & ResNet + Visual Dictionary   & Single stream & MLM+MVM+ITM & COCO+VG\\
VL-T5~\shortcite{cho2021unifying}     & T5, BART  & Faster R-CNN   & Single stream & \makecell[l]{MLM+VQA+ITM+VG+GC} & COCO+VG\\
XGPT~\shortcite{xia2021xgpt}     & transformer  & Faster R-CNN   & Single stream & \makecell[l]{IC+MLM+DAE+MRFR} & CC\\
Visual Parsing~\shortcite{xue2021probing} & BERT & Faster R-CNN + Swin transformer & Dual stream & MLM+ITM+MFR & COCO+VG\\
ALBEF~\shortcite{li2021align}    & BERT  & ViT   & Dual stream  & MLM+ITM+CMCL & CC+COCO+VG+SBU\\
SimVLM~\shortcite{wang2021simvlm}     & ViT  & ViT   & Single stream  & PrefixLM & C4+ALIGN\\
WenLan~\shortcite{huo2021wenlan}    & RoBERTa  & Faster R-CNN + EffcientNet   & Dual stream  & CMCL & RUC-CAS-WenLan\\
ViLT~\shortcite{kim2021vilt}    & ViT  & Linear Projection   & Single stream  & MLM+ITM & CC+COCO+VG+SBU\\
\midrule
\textbf{\small{Dual Encoder}} \\
CLIP~\shortcite{radford2021learning}  & GPT2  & ViT, ResNet   &   & CMCL & self-collected\\
ALIGN~\shortcite{jia2021scaling}    & BERT  & EffcientNet   &   & CMCL & self-collected\\
% DeCLIP\shortcite{li2021supervision} & GPT2, BERT & ViT, ResNet, RegNetY-64GF &   & CMCL+MLM+CL & CC+YFCC+self-collected \\
DeCLIP~\shortcite{li2021supervision} & GPT2, BERT & ViT, ResNet, RegNetY-64GF &   & CMCL+MLM+CL & CC+self-collected \\

\midrule
\textbf{\makecell[l]{\small{Fusion Encoder+ Dual Encoder}}} \\
VLMo~\shortcite{wang2021vlmo}  & BERT  & ViT   & Single stream & MLM+ITM+CMCL & CC+COCO+VG+SBU\\
% FLAVA\shortcite{singh2021flava}    & ViT  & ViT   & Single stream   & MMM+ITM+CMCL & CC+COCO+VG+SBU+RedCaps+Wiki+Localized~Narratives+YFCC100M\\
FLAVA~\shortcite{singh2021flava}    & ViT  & ViT   & Single stream   & MMM+ITM+CMCL & CC+COCO+VG+SBU+RedCaps\\
\bottomrule
\end{tabular}}
\caption{Glossary of Representative VL-PTMs. {MLM/MVM: (Cross-Modal) Masked Language/Vision Modeling. ITM: Image-Text Matching. MRC: Masked Region Classification. MRFR: Masked Region Feature Regression. VG: Visual Grounding. GC: Grounded Captioning. WRA: Word-Region Alignment. CMCL: Cross-Modal Contrastive Learning. DAE: Denoising AutoEncoding}} 
\label{summarize}
\end{table*}
\section{Learning Vision-Language Representation}\label{inputrep}
As discussed in Section~\ref{intro}, encoding images and texts as embeddings preserving input semantics is the first step in pre-training a VL-PTM. The ways of encoding images and texts are quite different because of the discrepancy between the two modalities. Almost all VL-PTMs utilize a transformer-based PTM as a text encoder, but how to learn visual representations based on visual contents is still an open problem. %We will explain why the connection between visual concepts is so crucial for V-L tasks in section~\ref{image-rep}.} 
In what follows, we introduce several methods to encode the images and texts into single-modal embeddings before feeding them into a cross-modal transformer.

% \paratitle{Pre-training Datasets.}~
% \textcolor{blue}{Data is the core driver of large VL-PTMs. Most of VL-PTMs use a large number of image-text pairs for pre-training. Some of the datasets are elaborate-labeled~(\emph{e.g.} MS-COCO) while some are crwaled from websites~(\emph{e.g.}LAION). In Table \ref{datasets} we list several widely-used or recently proposed pre-training datasets.}

\begin{table}
\small
\footnotesize
\begin{tabular}{lll}
\hline
Dataset  & Size & Reference \\
\hline
COCO       & 328,124  & \cite{lin2014microsoft}     \\
VG       & 108,077  & \cite{krishna2017visual}      \\
CC    & 3.1M  & \cite{sharma2018conceptual}     \\
SBU   & 1M  & \cite{ordonez2011im2text}     \\
LAION & 400M & \href{https://laion.ai/laion-400-open-dataset/}{https://laion.ai/laion-400-open-dataset/} \\ 
RedCaps & 12M & \cite{desai2021redcaps} \\

\hline
\end{tabular}
\caption{Widely Used Pre-training Datasets}
\label{datasets}
\end{table}

\paragraph{Pre-training Dataset.}The initial step in pre-training VL-PTMs is to construct large-scale image-text pairs. We formally define the pre-training dataset as $\mathcal{D}=\{\left(W,  V\right)\}_{i=1}^N$, where $W$ and $V$ denote the text and image, respectively, and $N$ is the number of image-text pairs. Specifically, each text will be tokenized as a sequence of tokens $W=\langle w_1,...,w_n \rangle$. Similarly, each image will be also transformed into a sequence of object features (or grid features, or patch features), denoted as $V=\langle v_1,...,v_m \rangle$. In Table \ref{datasets} we list several widely-used or recently proposed pre-training datasets.

\paragraph{Text Representation.}Most of existing studies on VL-PTMs follow BERT~\cite{devlin2018bert} to preprocess the raw text. The text sequence is first split into tokens and concatenated with ``[CLS]'' and ``[SEP]'' tokens, denoted as $W=\langle \text{[CLS]}, w_1,...,w_n, \text{[SEP]} \rangle$. Each token $w_j$ will be mapped to a word embedding. Besides, a positional embedding indicating the position and a segment embedding indicating the modality type are added with 
the word embedding to obtain the final embedding of $w_j$. %By feeding the summation of word embeddings, positional embeddings, and segment embeddings into BERT, we can obtain the final input text representation of ${W}$, denoted as $\hat{E}({W})=\langle  \hat{E}_w(\text{[CLS]}), \hat{E}_w(w_1),...,\hat{E}_w(w_j),...,\hat{E}_w(w_n) \rangle$.

%A positional embedding and a segment embedding are used to encode position information and indicate index of the word’s sentence if multiple exists.
%The final input text representation is $\langle  \hat{w}_1,...,\hat{w}_j,...,\hat{w}_n \rangle$. For each of the final representation at position i, it is conculated as:
% \begin{equation}
%     \hat{w} = LayerNorm(w_i + p_i +s_w)
% \end{equation}
% where $w_i$ indicates the original word embedding, $p_i$ indicates positional embedding at index i, $s_w$ indicates index of word's sentence and LayerNorm indicates the same normalization function as BERT.

\paragraph{Image Representation.}\label{image-rep}To align with the sequence of embeddings of the paired text, the image $V$ will also be represented as a sequence of embedding vectors. In this way, we can unify input representation as a sequence of embeddings for both modalities. Unlike relationships among words in text, relationships among visual concepts in images are critical for V-L tasks but difficult to capture. For example, in order to generate a description of an image, the model is expected to infer the complex relationships among various objects in the image. Therefore, many works elaborate different vision encoders to model these relationships and the attributes of objects. Early works like ViLBERT~\cite{lu2019vilbert} and LXMERT~\cite{tan2019lxmert} first utilize Faster R-CNN~\cite{ren2015faster} to detect a sequence of object regions from an image and then encode them as a sequence of Region-Of-Interest~(ROI) features. Besides, some VL-PTMs get rid of the bounding box and encode an image into pixel-level grid features. For example, pixel-BERT~\cite{huang2020pixel} and SOHO~\cite{huang2021seeing} abandon Faster R-CNN in favor of ResNet so that the visual encoder could view an image as a whole, avoiding the risk of neglecting some critical regions. Apart from these methods, many works try to follow the success of ViT~\cite{dosovitskiy2020image} to utilize a transformer to extract vision features. In this scenario, the transformer in VL-PTMs is tasked with the objective of modeling object relationships within an image. An image is firstly split into several flattened 2D patches. Then the embeddings of image patches are arranged in a sequence to represent the original image.
%For an image represented as $x \in \mathbb{R}^{H \times W \times C}$, it can be split into a sequence of flatten 2D patches $x_p \in \mathbb{R}^{M \times (P^2\cdot C)}$, where $(H, W)$ is the resolution of the original image, $C$ is the number of channels, $(P, P)$ is the size of each image patch, and $M$ is the resulting number of patches.
ALBEF~\cite{li2021align} and SimVLM~\cite{wang2021simvlm} feed patches to an ViT encoder to extract vision features, which lead the way to a full-transformer VL-PTM. %ViLT~\cite{kim2021vilt} takes a step further to throw away the ViT encoder while simply uses a linear projection layer to encode the patch sequence as image features.

\section{Modeling Vision-Language Interaction}\label{architecture}
After encoding images and texts into single-modal embeddings, the next step is to design an encoder to integrate information from both vision and language modalities. For example, to answer a question about an image, the model needs to combine the linguistic information from both questions and answers, then localize the corresponding region in the paired images, and lastly align linguistic meanings with visual clues. Based on the way of aggregating information from different modalities, we categorize the encoder into \textit{fusion encoder}, \textit{dual encoder} and the combination of both. %Borrowing ideas from NLP, the architectures of VLPs are quite similar to those of PLMs. The mainstream architectures can be divided into two categories, namely \textit{encoder-only} and \textit{encoder-decoder}, based on their ways of modeling the relationship between input and output. 

% \subsection{Encoder-only}
% The approaches of using a single encoder to model V-L interaction can be further divided into two categories: fusion encoder and dual encoder. \textcolor{blue}{MoE?}
\subsection{Fusion Encoder}
The fusion encoder takes text embeddings and image features as input and designs several fusion approaches to model V-L interaction. After self-attention or cross-attention operation, the hidden states of the last layer will be treated as the fused representation of different modalities. There are mainly two types of fusion schemes for modeling the cross-modal interaction: single stream and dual stream.

\paragraph{Single-stream Architecture.}The single-stream architecture assumes that the potential correlation and alignment between two modalities are simple, which can be learned by a single transformer encoder. Therefore, the text embeddings and image features are concatenated together, adding some special embeddings to indicate position and modalities, and fed into a transformer-based encoder. 

Although different V-L tasks require different input formats~(\emph{e.g.}, $\langle$\textit{caption}, \textit{image}$\rangle$ for image captioning, $\langle$\textit{question}, \textit{answer}, \textit{image}$\rangle$ for VQA), single-stream architecture can handle them in a unified framework due to the unordered representation nature of transformer attention. VisualBERT~\cite{li2019visualbert} and V-L BERT~\cite{su2019vl} utilize segment embedding to indicate input elements from different sources. Instead of simply using image-text pair, OSCAR~\cite{li2020oscar} adds object tags detected from the image and represents the image-text pair as a $\langle$\textit{word}, \textit{tag}, \textit{image}$\rangle$ triple to help the fusion encoder better align different modalities. Since the single-stream architecture performs self-attention directly on two modalities, they may neglect intra-modality interaction. Thus some works propose to employ dual-stream architecture to model V-L interaction. 

\paragraph{Dual-stream Architecture.}Different from self-attention operation in single-stream architectures, dual-stream architectures adopt a cross-attention mechanism to model V-L interaction, where the query vectors are from one modality while the key and value vectors are from the other. A cross-attention layer usually contains two unidirectional cross-attention sub-layers: one from language to vision and another from vision to language. They are responsible for exchanging information and aligning the semantics between the two modalities.

Dual-stream architectures assume that the intra-modal interaction and cross-modal interaction need to be separated to obtain better multimodal representations. ViLBERT~\cite{lu2019vilbert} utilizes two transformers to further model intra-modality interaction after the cross-modal module. LXMERT~\cite{tan2019lxmert} does not use extra transformers but appends a self-attention sub-layer after cross-attention sub-layer to further build internal connections. In the cross-modal sub-layers, the parameters of the attention module are shared between the two streams. In this case, the model learns a single function to contextualize image and text embeddings. ALBEF~\cite{li2021align} employs two separate transformers before cross-attention for images and texts, which better decouples intra-modal interaction and cross-modal interaction. This type of architecture helps to encode the input in a more comprehensive way. However, the extra feature encoder makes it parameter-inefficient.

\subsection{Dual Encoder}
Although the fusion encoder could model cross-modal interaction at different levels and achieves state-of-the-art results on many V-L tasks, it relies on a heavy transformer network to model V-L interaction. When performing cross-modal matching tasks like Image-Text Retrieval, the fusion encoder has to jointly encode all possible image text pairs, which leads to a quite slow inference speed.

In contrast, a dual encoder utilizes two single-modal encoders to encode two modalities separately. Then, it adopts straightforward methods such as shallow attention layer~\cite{lee2018stacked} or dot product~\cite{radford2021learning,jia2021scaling} to project the image embedding and text embedding to the same semantic space for computing V-L similarity scores. Without the complex cross-attention in transformer, the V-L interaction modeling strategy in the dual encoder is much more efficient. Thus, the feature vectors of images and text can be pre-computed and stored, which is more effective for retrieval tasks than the fusion encoder. Although dual encoder models like CLIP~\cite{radford2021learning} have shown surprising performance on image-text retrieval tasks, they fail in some hard V-L understanding tasks such as NLVR~\cite{suhr2018corpus}. This is attributed to the shallow interaction between the two modalities.  

\subsection{Combination of Fusion Encoder and Dual Encoder}
Based on the observation that fusion encoder performs better on V-L understanding tasks while dual encoder performs better on retrieval tasks, it is natural to combine the benefits of the two types of architectures. FLAVA~\cite{singh2021flava} first adopts a dual encoder to obtain single-modal representations. Then the single-modal embeddings are sent to a fusion encoder to obtain cross-modal representation. Apart from its model design, FLAVA conducts several unimodal pre-training tasks to improve the quality of single-modal representations. VLMo~\cite{wang2021vlmo} introduces \textbf{M}ixture-\textbf{o}f-\textbf{M}odality-\textbf{E}xpert~(\textbf{MoME}) and unifies a dual encoder and a fusion encoder into a single framework. After pre-training on images, texts, and image-text pairs by stage, VLMo can not only be fine-tuned on V-L understanding tasks, but also be applied to efficient image-text retrieval.
\section{Cross-Modal Pre-training Tasks}\label{pretraining}
According to Section~\ref{intro}, after the input images and texts are encoded as vectors and fully interacted, the next step is to design pre-training tasks for VL-PTMs. The designed pre-training tasks have a great impact on what VL-PTM can learn from the data. In this section, we introduce some widely-used pre-training tasks.

% \subsection{Language Tasks}
% After masking some tokens in the text, V-L PTMs take the corrupted image-text pair as input to predict the masked tokens. Based on the way of generating tokens, this pre-training task can be divided into Masked Language Modeling~(MLM) and Sequence-to-Sequence Language Modeling~(Seq2Seq LM).

\subsection{Cross-Modal Masked Language Modeling~(MLM)} 
Cross-modal MLM is similar to MLM in the BERT model. In cross-modal MLM, VL-PTMs predict masked tokens not only based on unmasked tokens, but also by taking vision features into account. The dependency on vision modality differentiates cross-modal MLM from MLM in NLP. This task has been proven to be quite effective for pre-training VL-PTMs because it helps the model to align vision and text by considering the relationship between image and text. Formally, the objective can be defined as:
\begin{equation}\small
    {L}_{\mathrm{MLM}}=-\mathbb{E}_{\left(W, V\right) \in \mathcal{D}}\log P_{\theta} \left(w_m|w_{\backslash m}, V \right),
\end{equation}
where $w_m, w_{\backslash m}$ represent the masked tokens and unmasked tokens respectively, and $\left(W, V\right) \in \mathcal{D}$ represents a text $W$ and an image $V$ sampled from dataset $\mathcal{D}$. 

Due to the distinction between cross-modal MLM and MLM in NLP, an effective masking strategy is necessary for cross-modal MLM. 
If the method is too simple, the model may be able to predict the masked tokens purely on the basis of their surrounding tokens. By masking some tokens that rely on the image, VL-PTMs will take into account the image features, thus aligning tokens and their corresponding objects in the image. ViLT~\cite{kim2021vilt} utilizes the Whole Word Masking strategy, which prevents the model from predicting tokens solely by words co-occurrence; InterBERT~\cite{lin2020interbert} masks several consecutive segments of text to make this pre-training task more difficult and improves its performance on downstream tasks further.

% \paratitle{Sequence-to-Sequence Language Modeling~(Seq2Seq LM)} is different from MLM such that it enables bi-directional attention in the prefix and generate the masked tokens auto-regressively. The objective can be formulated as:
% \begin{equation}
%   \mathcal{L}_{\mathrm{seq2seq}}=-\mathbb{E}_{\left(V,W\right) \in D}\sum_{i=T_p+1}^N P_{\theta}\left(x_i|x_{<i}, V\right), 
% \end{equation}
% where $T_p$ and $N$ represent the length of prefix and sequence respectively.

% Unified VLP~\cite{zhou2020unified} conditions on image features and generates the corresponding captions auto-regressively. For a given image-text pair, SimVLM~\cite{wang2021simvlm} prepends the image patch features to the text sequence, and samples a prefix from the text tokens as condition to generate the remaining ones.

\subsection{Cross-Modal Masked Region Prediction~(MRP)}
Similar to the cross-modal MLM, cross-modal MRP masks some RoI features with zeros and predicts them based on other image features. The model learns object relationships by inferring from other unmasked regions and learns V-L alignments by inferring from the text. There are two kinds of learning objectives: Masked Region Classification~(MRC) and Masked Region Feature Regression~(MRFR). 

\paragraph{Masked Region Classification~(MRC).}MRC learns to predict the semantic class of each masked region. This task is motivated by the observation that VL-PTMs just learn high-level semantics of images instead of raw pixels from the language side. To predict the region class, the hidden state $\boldsymbol{h}_{v_i}$ of the masked region $v_i$ from VL-PTMs is fed into a fully-connected~(FC) layer, followed by a softmax function to form a predicted distribution on $K$ object classes. The final objective is to minimize the cross-entropy (CE) loss between the predicted distribution and the detected object category, which can be formally defined as:
\begin{equation}\small
    \mathcal{L}_{\mathrm{MRC}}=\mathbb{E}_{\left(W, V\right) \in \mathcal{D}}\sum_{i=1}^{l}\mathrm{CE}\big(\mathrm{softmax}(\mathrm{FC}(\boldsymbol{h}_{v_i})),c(v_i)\big),
\end{equation}
where $l$ is the amount of masked regions, and $c\left(v_i\right)$ represents the true label of the masked region, such as the object detection output or the (pre-defined) visual tokens.

%LXMERT~\cite{tan2019lxmert}, VL-BERT~\cite{su2019vl}, Unicoder~\cite{li2020unicoder} and UNITER~\cite{chen2020uniter} use Faster R-CNN to extract region features for Masked Region Classification. 

Through the cross-modal attention in VL-PTMs, the hidden state $\boldsymbol{h}_{v_i}$ contains information from both vision and language modality, which makes it possible to predict visual semantic class from the text. As for the ground-truth label $c\left(v_i\right)$, an intuitive method is to regard object tags~(with the highest confidence score) detected from object dector as the true labels\cite{tan2019lxmert,su2019vl}. However, these labels are pseudo, which highly relies on the quality of the pre-trained object detectors, thus there are some variants of this task. ViLBERT~\cite{lu2019vilbert} and UNITER~\cite{chen2020uniter} propose to consider the raw output of the detector as soft labels, which is a distribution of object classes. In this scenario, the objective becomes the KL-divergence between two distributions. SOHO~\cite{huang2021seeing} first maps the CNN-based grid features to visual tokens, and then predicts the masked visual tokens based on their surrounding tokens.

\paragraph{Masked Region Feature Regression~(MRFR).}MRFR learns to regress the masked region feature $\boldsymbol{h}_{v_i}$ to its corresponding original region feature $\hat{E}_V(v_i)$, which can be written as:
\begin{equation}\small
   \mathcal{L}_{\mathrm{MRFR}}=\mathbb{E}_{\left(W, V\right) \in \mathcal{D}}\sum_{i=1}^l \Vert \mathrm{FC}(\boldsymbol{h}_{v_i})-\hat{E}_{V}(v_i) \Vert^2 .
\end{equation}

%The notations are the same as MRC, except $v_i$ could be various vector representations of visual region $i$~(\emph{e.g.} input embeddings, transformer hidden states). 
In this formula, the region feature $\hat{E}_V(v_i)$ of $v_i$ is computed based on an unmasked image and $l$ represents the amount of masked regions. MRFR requires the model to reconstruct the high-dimensional vectors instead of semantic class. When images are represented as a sequence of region features by faster R-CNN, simple masking strategies like random masking can give satisfying performances~\cite{tan2019lxmert,chen2020uniter,li2020unimo}. However, random masking will not be so effective when images are converted into grid features or patch features, because the model will directly duplicate neighbor features as the predicted features. Visual parsing~\cite{xue2021probing} uses patch features to represent an image and assumes that visual tokens~(region features) of high attention weights have similar semantics. It first randomly masks a visual token as a pivot token, and continues to mask $k$ tokens with top-$k$ attention weights. SOHO\cite{huang2021seeing} pre-trains a vision dictionary and masks all the features sharing the same visual index to avoid information leakage.

% \subsection{Cross-Modal Learning}
% Language and visual learning aim to learn fine-grained correlation between images and texts, while cross-modal learning aims to learn coarse-grained alignment between them.

\subsection{Image-Text Matching~(ITM)}
Cross-modal MLM and MRP help VL-PTMs learn the fine-grained correlation between images and texts, while ITM empowers VL-PTMs with the ability to align them at a coarse-grained level. ITM is similar to the Next Sentence Prediction~(NSP) task in NLP, which requires the model to determine whether an image and a text are matched. Given an image-text pair, a score function $s_{\theta}$ measures the alignment probability between the image and text. The objective function is:
\begin{equation}\small
    \begin{aligned}
        \mathcal{L}_{\mathrm{ITM}}=&-\mathbb{E}_{\left(W, V\right) \in \mathcal{D}}\left[y \log s_{\theta}\left(\boldsymbol{h}_{w_{[\text{CLS}]}}, \boldsymbol{h}_{v_{[\text{IMG}]}}\right)\right.\\
        &\left.+(1-y) \log \left(1-s_{\theta}\left(\boldsymbol{h}_{w_{[\text{CLS}]}}, \boldsymbol{h}_{v_{[\text{IMG}]}}\right)\right)\right],
    \end{aligned}
\end{equation}
where $y \in \{0,1\}$ represents whether $W$ and $V$ are matched with each other or not, and $\boldsymbol{h}_{w_{[\text{CLS}]}}$ and $\boldsymbol{h}_{v_{[\text{IMG}]}}$ are the representations of $w_{\text{[CLS]}}$ and $v_{\text{[IMG]}}$, respectively. 

The key to this task is how to represent an image-text pair in a single vector so that the score function $s_{\theta}$ could output a probability. UNITER~\cite{chen2020uniter}, Unicoder~\cite{li2020unicoder} and SOHO~\cite{huang2021seeing} 
concatenate the word sequence $W$ and the object sequence $V$ and take the final hidden state of the ``[CLS]'' token as the fused representation. By feeding it into a fully-connected layer layer, they can reduce the dimension to predict the alignment probability. While, ViLBERT~\cite{lu2019vilbert} adopts the representation of the ``[IMG]'' and ``[CLS]'' tokens to represent image and text respectively, and the fused representation is computed by element-wise product between them.

\subsection{Cross-Modal Contrastive Learning~(CMCL)}
CMCL aims to learn universal vision and language representation under the same semantic space by pushing the embeddings of matched image-text pairs together while pushing the non-matched ones apart. The image-to-text contrastive loss can be formulated as:
\begin{equation}\small
    \mathcal{L}_{\mathrm{i2t}}=-\mathbb{E}_{\left(W, V\right) \in \mathcal{D}}\left[\log \frac{s_{\theta} \left(\boldsymbol{h}_{v_{[\text{IMG}]}}, \boldsymbol{h}_{w_{[\text{CLS}]}}\right)}{\sum_{W'} s_{\theta} \left(\boldsymbol{h}_{v_{[\text{IMG}]}}, \boldsymbol{h}_{w'_{[\text{CLS}]}}\right)}\right],
\end{equation}
where $W'$ belongs to the negative samples set of $V$, $\boldsymbol{h}_{w_{[\text{CLS}]}}$, $\boldsymbol{h}_{v_{[\text{IMG}]}}$ and $\boldsymbol{h}_{w'_{[\text{CLS}]}}$ are the representations of $w_{\text{[CLS]}}$, $v_{\text{[IMG]}}$ and $w'_{\text{[CLS]}}$, respectively, and $s_{\theta}$ is a score function to justify how similar a given image-text pair is. It is worth noting that the contrastive loss in CMCL is symmetrical, and the text-to-image contrastive loss is formulated similarly.

CLIP~\cite{radford2021learning} and ALIGN~\cite{jia2021scaling} leverage large-scale image-text pairs to learn transferable visual representations and exhibit surprising zero-shot transfer to image classification tasks. ALBEF~\cite{li2021align} proposes to adopt momentum distillation to facilitate contrastive learning on massive noisy image-text pairs. WenLan~\cite{huo2021wenlan} employs MoCo~\cite{he2020momentum} and maintains a queue to store negative samples, which has been proven to be effective for contrastive learning. UNIMO~\cite{li2020unimo} incorporates a large volume of unimodal data during contrastive learning, allowing vision and language to enhance each other. It outperforms previous works on both multimodal and unimodal downstream tasks. \cite{yang2022vision} claims that CMCL does not guarantee similar inputs from the same modality stay close by, so they introduce intra-modal contrastive learning to benefit representation learning.

% \paratitle{Incorporating Downstream Tasks} as pre-training tasks can facilitate V-L PTMs  better adapt to downstream tasks. LXMERT~\cite{tan2019lxmert} finds that image QA leads to a better Cross-Modal representation. \citeauthor{cho2021unifying} incorporates VQA and visual grounding as pre-training tasks. XGPT~\cite{xia2021xgpt} uses image captioning as pre-training task so that the pre-trained model can be directly fine-tuned on downstream image captioning task.

%\subsection{Discussion}
\section{Adapting VL-PTMs to Vision-Language Downstream Tasks}\label{downstream}
Pre-training tasks are able to help VL-PTMs to learn general visual and linguistic features, which can be applied to various downstream tasks. In this section, we introduce several common vision-language integration tasks and how VL-PTMs are adapted to them. Basically, we categorised these downstream tasks into cross-modal matching, cross-modal reasoning and vision and language generation. %As an overview, we list all the tasks and their evaluation metrics in table \ref{metrics}.

% \begin{table}
% \footnotesize
% \centering
% \begin{tabular}{lll}
% \hline
% Task  & Metrics  \\
% \hline
% ITR       & Recall@K   \\
% VRE   & Accuracy, Precision  \\
% %VQA    & Accuracy     \\
% %NLVR   & Accuracy    \\
% %VCR & Accuracy  \\ 
% %VE & Accuracy  \\
% VQA, NLVR, VR, VE    & Accuracy     \\

% IG & \makecell[l]{Inception score, Inception v3\\Frechet Inc\'eption Distance} \\
% IC & \makecell[l]{BLEU, METEOR\\ CIDEr, SPICE}\\
% MMT & BLEU, METEOR\\
% %Visual Dialog & \makecell[l]{Recall@K\\ MR, MRR\cite{das2017visual}}\\

% \hline
% \end{tabular}
% \caption{Tasks and Metrics}
% \label{metrics}
% \end{table}

\subsection{Cross-Modal Matching} % Cross Modal Matching?
Cross-modal matching requires VL-PTMs to learn cross-modal correspondences between different modalities. We introduce two commonly-used cross-modal matching tasks: image text retrieval and visual referring expression.

\paragraph{Image Text Retrieval~(ITR).}ITR is a typical cross-modal matching task. This task requires retrieving an image that matches a given sentence most and vice versa. Early VL-PTMs that utilize a fusion-encoder architecture obtain a fused vector representation which is later projected to a similarity score~\cite{lu2019vilbert,li2019visualbert,li2020oscar}. %The computation process requires jointly encoding all the image-text pairs and gives a quadratic time complexity. 
Dual-encoder architectures such as CLIP~\cite{radford2021learning} and ALBEF~\cite{li2021align} are more efficient for ITR as they can pre-compute and store the embeddings of images and texts before retrieval. %The matching score is measured by the similarity of the two projections, giving a linear time complexity and a much faster inference speed.

\paragraph{Visual Referring Expression~(VRE).}VRE is an extension of the referring expression task in NLP. The goal is to localize the region in an image that corresponds to a specific textual description. Most VL-PTMs~(\emph{e.g.}~\cite{lu2019vilbert}) take the final representation of the extracted region proposals as input and learn a linear projection to predict a matching score, which is the same strategy as ITR during fine-tuning. 

\subsection{Cross-Modal Reasoning} 
Cross-modal reasoning requires VL-PTMs to perform language reasoning based on visual information. Ignoring any modality gives poor performance. Here we present two commonly-used cross-modal reasoning tasks.

\paragraph{Visual Question Answering~(VQA).}VQA is a widely-used cross-modal reasoning task. Different from text-based QA, VQA requires answering questions about images. Most researchers consider VQA as a classification task and require the model to select a correct answer from an answer pool. VL-PTMs with a fusion-encoder architecture usually map the final cross-modal representation~(usually corresponds to the input [CLS] token) to the distribution of answer labels. However, VL-PTMs with a dual-encoder architecture are not so effective for VQA tasks because the interaction between the two modalities is too shallow to conduct cross-modal reasoning. There are also some works modeling VQA as a generation task~\cite{cho2021unifying,wang2021simvlm}, which can generalize better to real-world open-ended scenarios. %The VQA task is relatively harder for dual-encoder-based VL-PTMs as they don't output a fused representation. CLIP-VIL\cite{shen2021much} uses other traditional methods and considers CLIP architecture simply as a visual encoder. CLIP-VIL doesn't achieve a promising result on the VQA task as on the Image Text Retrieval task, which indicates that dual-encoder architecture may not be suitable on Cross Vision and Language Reasoning task.

\paragraph{Natural Language for Visual Reasoning~(NLVR).}NLVR provides an image pair and a textual statement as input and requires the model to decide whether the statement is true about the image pair, thus can be considered as a binary classification task. Most VL-PTMs first encode the given two image-text pairs separately, then a classifier is trained over the concatenation of the two embeddings to make a prediction~\cite{tan2019lxmert,chen2020uniter}.
%Most VL-PTMs concatenate given image pair and textual statement as input and use the same approach in VQA task to solve this problem.~\cite{tan2019lxmert,chen2020uniter}.

\paragraph{Visual Commonsense Reasoning~(VCR).}VCR is considered to be another kind of VQA task. The main difference between VCR and VQA is that VCR's questions pay more attention to visual common sense. Different from VQA, the VCR task can be decomposed into two multi-choice sub-tasks: question answering (Q → A) and answer justification (Q + A → R). Most VL-PTMs utilize the same approach in VQA to solve these two sub-tasks. For the question answering sub-task, the procedure is the same as VQA. For the answer justification sub-task, the concatenations of question and answer are treated as the new questions and the rationales become the options. A linear layer is
trained to predict a score for each possible option~\cite{lu2019vilbert}.%Most VL-PTMs use the concatenation of question tokens, answer tokens and rational tokens as input and treat VCR as a classification task like VQA~\cite{lu2019vilbert}. Different from the VQA dataset, VCR integrates object tags into the language providing direct grounding supervision.

%\paratitle{Visual Entailment(VE)} is a visual reasoning task to predict whether the relationship between an image and a text is entailment, neutral, or contradictory. Still, solution for this task is to take certain final representations and use a linear projection to predict an answer.

\subsection{Vision and Language Generation}
Based on the source modal and target modal, the generation task can be divided into text-to-image generation and image-to-text generation~(multimodal text generation).

\paragraph{Text-to-Image Generation.}Text-to-Image generation is the task of generating a corresponding image from a descriptive text.
X-LXMERT~\cite{cho2020x} first converts continuous visual representations to discrete cluster centroids and then ask the model to predict the cluster ids of masked regions. DALL-E~\cite{ramesh2021zero} trains a codebook to tokenize images and formulates text-to-image generation task as an autoregressive generative task. It achieves new state-of-the-art results on MS-COCO~\cite{lin2014microsoft} in zero-shot setting.

\paragraph{Multimodal Text Generation.}Multimodal text generation can be regarded as a special type of conditional text generation, where the condition includes not only texts but also images. Usually, a decoder is needed for the generation process. Image captioning is a typical image-to-text generation task that requires the model to generate a description of an image. XGPT~\cite{xia2021xgpt} and VL-T5~\cite{cho2021unifying} encode the images first and then employ a decoder to generate the captions autoregressively. Multimodal machine translation is another generation task that aims to introduce images to improve translation quality. VL-T5~\cite{cho2021unifying} tackles this task using the same strategy as in image captioning. %Visual dialog is first proposed as a new task by \cite{das2017visual}. Specifically, given an image, a dialog history, and a question about the image, the system is required to answer the question according to the image. \cite{murahari2020large} first adapts VL-PTM, ViLBERT~\cite{lu2019vilbert} to this task, which is pre-trained on CC~\cite{sharma2018conceptual} and VQA~\cite{antol2015vqa}, and fine-tuned on VisDial~\cite{das2017visual}.

As for the connection between VL-PTMs architecture and downstream tasks, fusion encoder is more suitable than dual encoder on cross-modal reasoning tasks for its powerful ability to model interaction. The dual encoder is more suitable for cross-modal retrieval tasks since it keeps a similar performance as fusion encoder does while being more efficient.

\section{Conclusion and Future Directions}\label{conclusion}
In this paper, we present an overview of VL-PTMs. We review the commonly-used architectures and discuss their advantages and disadvantages. We also introduce several mainstream approaches to pre-training a VL-PTM and adapt it to downstream tasks. Though VL-PTMs have made significant progress on V-L tasks compared to traditional methods, there are still several challenges that could be the directions of future research.

\paragraph{Unified Model Architecture.}Transformer-based models have shown surprising performance on NLP, CV, and multimodal tasks. The success of transformer-based models in various domains indicates the possibility of using a single transformer model to learn a representation of different modalities and building a general agent to handle tasks in different domains. UNIMO~\cite{li2020unimo} and FLAVA~\cite{singh2021flava} make some
inspiring attempts in this direction, but their performance on some tasks is much worse than the task-specific baselines. Data2vec~\cite{baevski2022data2vec} adopts self-supervised learning to unify vision, speech , and language. This model achieves competitive results to predominant methods on several tasks, which paves the way for a powerful unified model.

\paragraph{Model Compression and Acceleration.}Despite the great success achieved by VL-PTMs in various fields, it is difficult to deploy such a huge model in real-life scenarios, thus leading to a direction of VL-PTM compression and acceleration. Knowledge distillation has been used to compress VL-PTM~\cite{fang2021compressing}, but some other traditional compression methods such as quantization and pruning for VL-PTMs are yet to be explored. As for model acceleration, \citet{li2021supervision} construct a data-efficient paradigm for V-L pre-training. Despite all these achievements, only a few efforts focus on improving VL-PTM's inference speed.%Although these models prove to be effective, there are still a lot of methods to be explored.

\paragraph{Advanced Pre-training Methods.}Though the current pre-training method seems quite effective, the potential of advanced pre-training methods is yet to be explored. Using adversarial samples to enhance pre-training has been shown to be effective~\cite{gan2020large}, which helps VL-PTMs to overcome the overfitting issue. Stage-wise pre-training~\cite{wang2021vlmo} has been proposed for better single-modal representation. With that ahead, the potential of pre-training methods are not fully developed, which is worth further studies.%Some works have proposed to tackle such problems. \cite{gan2020large} uses adversarial training strategy to enhance VL-PTM's robustness, which is shown to be effective. But there still lacks of studies of new pre-training methods.

%The vision-language pre-trained model is always hungry for data, which leads to a concern of overfitting when fine-tuning on downstream tasks. Using adversarial training strategy to enhance model robustness has been shown to be effective~\cite{gan2020large}, and it is promising to investigate how to build a robust VL-PTM. %Also, the commonly-used dataset for pre-training is relatively clean, while most of the image-text pairs crawled from the internet can be very noisy. It is still a challenging task to build a robust model that can work well under such scenario.

\paragraph{Reaching the Limit of VL-PTMs.}Nowadays, with the success of large-scale PLMs in NLP, many researchers have also tried to build a deeper model or use a larger dataset for V-L pre-training. ALIGN~\cite{jia2021scaling} has a number of 675.4 million parameters and collects a huge dataset consisting of 1.8 billion image-text pairs for pre-training. It achieves state-of-the-art results on almost all downstream tasks. Wenlan~\cite{fei2021wenlan} expands the dataset to 650 million image-text pairs and shows astonishing performance on both vision-language understanding and generation tasks. In the future, VL-PTMs will need more high-quality data and more parameters to reach a higher recognition level.

\appendix

\bibliographystyle{named}
\bibliography{ijcai22}

\end{document}